\documentclass[review]{elsarticle}

\usepackage{amsmath}
\usepackage{hyperref} 
\usepackage{amsfonts}
\usepackage{amssymb,array}
\usepackage{caption}
\usepackage{subcaption}
\usepackage{xcolor}
\usepackage{graphicx}
\journal{Neurocomputing}

\newtheorem{remark}{Remark}
\newtheorem{notation}{Notation}
\newcommand{\prior}{p}

\begin{document}

\begin{frontmatter}


\title{Exact ICL maximization in a non-stationary temporal extension of the stochastic block model for dynamic networks}


\author[mymainaddress]{Marco Corneli}
\author[mymainaddress]{Pierre Latouche}
\author[mymainaddress]{Fabrice Rossi}


\address[mymainaddress]{Universit\'e Paris 1 Panth\'eon-Sorbonne - Laboratoire SAMM \\
90 rue de Tolbiac, F-75634 Paris Cedex 13 - France}

\begin{abstract}
  The stochastic block model (SBM) is a flexible probabilistic tool that can
  be used to model interactions between clusters of nodes in a
  network. However, it does not account for interactions of time varying
  intensity between clusters. The extension of the SBM developed in this paper
  addresses this shortcoming through a temporal partition: assuming
  interactions between nodes are recorded on fixed-length time intervals, the
  inference procedure associated with the model we propose allows to cluster simultaneously the nodes of the network and the time
  intervals.  The number of clusters of nodes and of time intervals, as well
  as the memberships to clusters, are obtained by maximizing an exact 
  integrated complete-data likelihood, relying on a greedy search
  approach. Experiments on simulated and real data are carried out in order to
  assess the proposed methodology.
\end{abstract}


\end{frontmatter}


\section{Introduction}\label{sec:intro}
Network analysis has been applied since the 30s to many scientific
fields. Indeed graph based modelling has been used in social sciences since
the pioneer work of Jacob Moreno \cite{Moreno}. Nowadays, network analyses
are used for instance in physics \citep{Albert2002}, economics \cite{1979},
biology \cite{barabasi2004, Palla2005} and history \cite{villa2008mining},
among other fields. 

One of the main tools of network analysis is clustering which aims at detecting
clusters of nodes sharing similar connectivity patterns. Most of the
clustering techniques look for \emph{communities}, a pattern in which nodes of
a given cluster are more likely to connect to members of the same cluster than
to members of other clusters (see \cite{FortunatoSurveyGraphs2010} for a
survey). Those methods usually rely on the maximization of the
\emph{modularity}, a quality measure proposed by Girvan and
Newman \cite{girvan2002community}. However, maximizing the modularity has been 
shown to be asymptotically biased \cite{bickel2009nonparametric}. 

In a probabilistic perspective, the stochastic block model (SBM)
\cite{Hollands83} assumes that nodes of a graph belong to hidden clusters and
probabilities of interactions between nodes depend only on these clusters. The
SBM can characterize the presence of communities but also more complicated
patterns \cite{inbooklatouche2009}. Many inference procedures have been
derived for the SBM such as variational expectation maximization (VEM)
\cite{daudin2008mixture}, variational Bayes EM (VBEM)
\cite{articlelatouche2012}, Gibbs sampling \cite{nowicki2001estimation},
allocation sampler \cite{mcdaid11}, greedy search \cite{Come_Latouche15} and non parametric schemes
\cite{kemp2006learning}. A detailed survey on the statistical and
probabilistic take on network analysis can be found in
\cite{Goldenberg09}.
   
While the original SBM was developed for static networks, extensions have been
proposed recently to deal with dynamic graphs. In this context, both nodes
memberships to a cluster and interactions between nodes can be seen as
stochastic processes. For instance, in the model of Yang et
al. \citep{yang2011detecting}, the connectivity pattern between clusters is
fixed through time and a hidden Markov model is used to describe cluster
evolution: the cluster of a node at time $t+1$ is obtained from its cluster at
time $t$ via a Markov chain. Conversely, Xu \emph{et al.}
\cite{xu2013dynamic} as well as Xing \emph{et al.}  \citep{xing2010} used a
state space model to describe temporal changes at the level of the
connectivity pattern. In the latter, the authors developed a method to
retrieve overlapping clusters through time.  

Other temporal variations of the SBM have been proposed. They generally share
with the ones described above a major assumption: the data set consists in a
sequence of graphs. This is by far the most common setting for dynamic
networks. Some papers remove those assumptions by considering continuous
time models in which edges occur at specific instants (for instance when
someone sends an email). This is the case of e.g. \cite{proceedingsdubois2013}
and of \cite{Rossi12,GuigouresEtAl2015}. The model developed in the present
paper introduces a sequence of graphs as an explicit aggregated view of a
continuous time model.

More precisely, our model, that we call the temporal SBM (TSBM), assumes that nodes belong to clusters that do not
change over time but that interaction patterns between those clusters have a
time varying structure. The time interval over which interactions are studied
is first segmented into sub-intervals of fixed identical duration. The model
assumes that those sub-intervals can be clustered into classes of homogeneous
interaction patterns: the distribution of the number of interactions that take
place between nodes of two given clusters during a sub-interval depends only on
the clusters of the nodes and on the cluster of the sub-interval. This
provides a non stationary extension of the SBM, which is based on the
simultaneous modelling of clusters of nodes and of sub-intervals of the time horizon.
Notice that a related approach is adopted in \citep{Come_Govaert14}, but with
a substantial difference: they consider time intervals whose membership is
known and hence exogenous, whereas in this paper the membership of
each interval is hidden and therefore inferred from the data.

The greedy search strategy proposed for the (original) stationary SBM 
 was compared with other SBM inference tools in many scenarios
 using both simulated and real data in  \cite{Come_Latouche15}.  Experimental results 
 emerged illustrating the capacity of  the method to retrieve relevant
 clusters. Note that the same framework was considered for the (related) latent block model
 \cite{wyse2014inferring},      in      the      context      of
 biclustering, and similar conclusions were drawn.  Indeed, contrary to most other techniques, this approach
 relies  on  an  exact  likelihood  criterion,  so  called  integrated
 complete-data likelihood (ICL), for optimization.  In
 particular,     it    does     not     involve    any     variational
 approximations. Moreover, it  allows the clustering of  the nodes and
 the  estimation   of  the   number  of   clusters  to   be  performed
 simultaneously. Alternative strategies usually
 do first the clustering for various number of clusters, by maximizing
 a given criterion, typically a lower
 bound. Then, they rely on a model selection criterion to estimate the
 number of clusters (see  \cite{daudin2008mixture} for instance). Some
 sampling   strategies  also   allow   the  simultaneous   estimation
 \cite{kemp2006learning,mcdaid11}.  However, the corresponding Markov
 chains  tend  to  exhibit  poor mixing  properties,  \emph{i.e.}  low
 acceptance  rates, for  large networks.   Finally, the  greedy search
 incurs \cite{Come_Latouche15} a smaller computational cost than existing techniques. Therefore, we
 follow the greedy  search approach and derive  an inference algorithm,
 for the new model we propose, which estimates the number of clusters, for both nodes and time intervals, as well as
memberships to clusters. 

{Finally, we cite the recent work of Matias et al. \cite{Matias_Poisson} who independently developed a temporal stochastic block model, related to the one proposed in this paper. Interactions in continuous time are counted by non homogeneous Poisson processes whose intensity functions only depend on the nodes clusters. A variational EM algorithm was derived to maximize an approximation of the likelihood and non parametric estimates of the intensity functions are provided.} 

This paper is structured as
follows: Section \ref{sec:S1} presents the proposed temporal extension of
the SBM and derives the exact ICL for this model. Section
\ref{sec:greedy_search} presents the greedy search algorithm used to maximize
the ICL. Section \ref{sec:experiments} gathers experimental results on
simulated data and on real world data. 

\section{A non stationary stochastic block model}\label{sec:S1}
We describe in this section the proposed extension of the stochastic block
model  (SBM)  to  non  stationary situations.  First,  we  recall  the
standard modeling assumptions of the SBM, then
introduce our temporal extension and finally derive an exact integrated
classification likelihood (ICL) for this extension. 

\subsection{Stochastic block model}
We consider a set of $N$ nodes $A=\{a_1, \dots, a_{N} \}$ and the $N\times N$
adjacency matrix $X=\{X_{ij}\}_{1\leq i,j \leq N}$ such that $X_{ij}$ counts
the number of direct interactions from $a_i$ to $a_j$ over the time interval
$[0,T]$. Self loops are not considered here, so the diagonal of $X$ is made of
zeros ($\forall i,\, X_{ii}=0$). Nodes in $A$ are assumed to belong to $K$ disjoint clusters 
\begin{equation*}
 A= \cup_{k \leq K}A_k, \qquad A_l \cap A_g=\emptyset, \quad\forall l\neq g.  
\end{equation*}
We introduce a hidden random vector $\mathbf{c}=\{c_1,\dots, c_N\}$, labelling each node's membership $c_i$
\begin{equation*}
 c_i=k\qquad\text{iff}\quad{i \in A_k,\quad \forall k\leq K}.
\end{equation*}
The $(c_i)_{1\leq i\leq N}$ are assumed to be independent and identically
distributed random variables with a multinomial probability distribution
depending on a common parameter $\boldsymbol{\omega}$ 
        \begin{equation*}
           \mathbf{P}\{ c_i=k \}=\omega_k \qquad\text{with}\qquad \sum_{k \leq K} \omega_k=1.
        \end{equation*}
Thus, node $i$ belongs to cluster $k$ with probability $\omega_k$.
As a consequence, the joint probability of vector $\mathbf{c}$ is
 \begin{equation}
 \label{eq:clabel}
  p(\mathbf{c} | \boldsymbol{\omega},K)=\prod_{k \leq K}\omega_k^{|A_k|},
 \end{equation}
where $|A_k|$ denotes the number of nodes in cluster $k$ (we denote $|U|$ the
cardinal of a set $U$). 

The first assumption of the original (stationary) SBM is that
interactions between nodes are independent given the cluster membership vector
$\mathbf{c}$, that is
\[
p(X|\mathbf{c})=\prod_{1\leq i, j\leq N}p(X_{ij}|\mathbf{c}).
\]
In addition, $X_{ij}$ is assumed to depend only on $c_i$ and $c_j$. More
precisely, let us introduce a $K\times K$ matrix  of model parameters
\begin{equation*}
\Lambda=\{\lambda_{kg}\}_{k\leq K, g \leq K}.
\end{equation*}  
Then, if $\mathbf{c}$ is such that $c_i=k$ and $c_j=g$, we assume that $X_{ij}$
is such that
\begin{equation*}
 p(X_{ij}| \mathbf{c}, \Lambda, K)=p(X_{ij}| \lambda_{kg}).
\end{equation*}
Combining the two assumptions, the probability of observing the adjacency
matrix $X$,
conditionally to $\mathbf{c}$, is given by
\begin{equation*}
 p(X| \mathbf{c}, \Lambda, K)=\prod_{k\leq K}\prod_{g \leq K}\prod_{i:c_i=k}\prod_{j:c_j=g}p(X_{ij}|\lambda_{kg}).
 \end{equation*}
When $X_{ij}$ characterizes interaction counts, a  common choice for $p(X_{ij}|\lambda_{kg})$ is the Poisson distribution.

\subsection{A non stationary approach}
In order to introduce a temporal structure, we modify the model described in
the previous section. The main idea is to allow interaction counts to follow
different regimes through time. The model assumes that interaction counts are
stationary at some minimal time resolution. This resolution is modeled via a
decomposition of the time interval $[0,T]$ in $U$ sub-intervals $I_u:=]t_{u-1}, t_{u}]$ delimited by the
following instants
\begin{equation*}
0=t_0 < t_1 < \dots < t_U=T,
\end{equation*}
whose increments
\begin{equation*}
t_u - t_{u-1}, \qquad u\in\{1,\dots, U\},
\end{equation*}
have all the same fixed value denoted $\Delta$. 

As for the nodes, a partition $C_1, \dots, C_D$ is considered for the time
sub-intervals. Thus, each $I_u$ is assumed  to belong to one of $D$ hidden
clusters and the random vector $\mathbf{y}=\{y_u\}_{u \leq U}$ is such that
\begin{equation*}
y_u=d \qquad\text{iff}\qquad I_u \in C_d, \forall d \leq D. 
\end{equation*}
A similar multinomial
distribution as the one of $\mathbf{c}$, is used to model $\mathbf{y}$ that is
\begin{equation}
\label{eq:ylabel}
 p(\mathbf{y} | \boldsymbol{\beta},D)=\prod_{d \leq D}\beta_d^{|C_d|},
\end{equation}
where $|C_d|$ is the cardinal of cluster $C_d$ and $\mathbf{P}\{y_u=d\}=\beta_d$. 

We now define $N^{I_u}_{ij}$ as the number of observed interactions from $i$
to $j$, in the time interval $I_u$. With the notations above, we have 
\begin{equation*}
 X_{ij}=\sum_{u=1}^{U}N_{ij}^{I_u}.
\end{equation*}
Following the SBM case, we assume conditional independence between all the
$N^{I_u}_{ij}$ given the two hidden vectors $\mathbf{c}$ and
$\mathbf{y}$. Denoting $N^{\Delta}=(N^{I_u}_{ij})_{1\leq i,j\leq N,1\leq u\leq
  U}$,  the  three  dimensional  tensor of  interaction  counts,  this
translates into
\[
p(N^{\Delta}|\mathbf{c},\mathbf{y})=\prod_{1\leq i,j\leq N,1\leq u\leq  U}p(N^{I_u}_{ij}|\mathbf{c},\mathbf{y}).
\]
Given a three dimensional $K \times K \times D$ tensor of parameters
$\Lambda=\{\lambda_{kgd}\}_{k\leq K, g \leq K, d\leq D}$, we assume that when
$\mathbf{c}$ is such that $c_i=k$ and $c_j=g$, and $\mathbf{y}$ is such that
$y_u=d$, then
\begin{equation*}
 p(N^{I_u}_{ij}| c_i=k, c_j=g, y_u=d)=p(N_{ij}^{I_u}|\lambda_{kgd}).
\end{equation*}
In addition, $N_{ij}^{I_u}|\lambda_{kgd}$ is assumed to be a Poisson
distributed random variable, that is
\begin{equation}
\label{eq:conditional_density}
 p(N_{ij}^{I_u}|\lambda_{kgd})=\frac{(\lambda_{kgd})^{N_{ij}^{I_u}}}{N^{I_u}_{ij}!}e^{-\lambda_{kgd}}.
\end{equation}

\begin{remark}
In the standard SBM, the adjacency matrix $X$ is a classical $N\times N$ matrix
and the parameter matrix $\Lambda$ is also a classical $K\times K$ matrix. In the
proposed extension, those matrices are replaced by three dimensional tensors,
$N^{\Delta}$ with dimensions $N\times N\times U$ and $\Lambda$ with dimensions
$K \times K \times D$. 
\end{remark}

\begin{remark}
  For $i$ and $j$ fixed and $\mathbf{c}$ known, the random variables
  $(N^{I_u}_{ij})_{1\leq u\leq U}$ are independent but are not identically
  distributed. As $u$ corresponds to time this induces a \emph{non stationary}
  structure as an extension of the traditional SBM. 
\end{remark}

\begin{notation}
To simplify the rest of the paper, let us denote
\begin{equation*}
\prod_{k,g,d} := \prod_{k \leq K} \prod_{g \leq K}\prod_{ d \leq D} \quad\text{and}\quad \prod_{c_i=k}:= \prod_{i:c_i=k}
\end{equation*}
and similarly for $\prod_{c_j=g}$ and $\prod_{y_u=d}$.  
\end{notation}
As in the case of the SBM, the distribution of $N^{\Delta}$, conditional to
$\mathbf{c}$ and $\mathbf{y}$, can be computed explicitly
\begin{align}
\label{eq:obsLL}
 p(N^{\Delta}|\Lambda, \mathbf{c}, \mathbf{y}, K,D)=&\prod_{k,g,d}\prod_{c_i=k}\prod_{c_j=g}\prod_{y_u=d} p(N_{ij}^{I_u}|\lambda_{kgd}),\nonumber \\
 =& \prod_{k,g,d} \frac{(\lambda_{kgd})^{S_{kgd}}}{P_{kgd}} e^{-\lambda_{kgd} R_{kgd}},
 \end{align}
where 
\begin{align*}
\label{eq:ICLNI}
 S_{kgd}&:= \sum_{c_i=k}\sum_{c_j=g} \sum_{y_u=d} N^{I_u}_{ij},\\
 P_{kgd}&:=\prod_{c_i=k} \prod_{c_j=g}\prod_{y_u=d} N_{ij}^{I_u}!, \\
 R_{kgd}&:=
 \begin{cases}
 |A_k||A_g||C_d|         & \quad\text{if}\quad g\neq k, \\
 |A_k|(|A_k|-1)|C_d|     & \quad\text{if}\quad g=k.
\end{cases}
\end{align*}

The full generative model is obtained by adding an independence assumptions
between $\mathbf{c}$ and $\mathbf{y}$ which gives to those vectors the
following joint distribution (obtained using equations \eqref{eq:clabel} and \eqref{eq:ylabel})
 \begin{equation}
 \label{eq:term_1_ns}
   p(\mathbf{c}, \mathbf{y} |\Phi, K, D) = \left(\prod_{k \leq K}\omega_k^{|A_k|}\right)\left(\prod_{d \leq D}\beta_d^{|C_d|}\right),
\end{equation}
where $\Phi=\{\boldsymbol{\omega},\boldsymbol{\beta} \}$.

{The identifiability of the proposed model could be assessed in future works, being outside the scope of the present paper. For a detailed and more general survey of the identifiability of the model parameters, in dynamic stochastic block models, the reader is referred to \citep{matias_dynamic_SBM}.}
\subsection{Exact ICL for non stationary SBM} 
The assumptions we have made so far are conditional on the number of clusters
$K$ and $D$ being known, which is not the case in real applications. A
standard solution to estimate the labels $\mathbf{c}$ and $\mathbf{y}$ as well
as the number of clusters would consist in fixing the values of $K$ and $D$ at
first and then in estimating the labels through one of the methods mentioned in the
introduction (e.g. variational EM). A model selection criterion could finally be used to choose the
values of $K$ and $D$. Many model selection criteria exist, such as the Akaike
Information Criterion (AIC) \cite{articleakaike1974}, the Bayesian Information
Criterion (BIC) \cite{Schwarz} and the integrated classification likelihood
(ICL), introduced in the context of Gaussian mixture models by Biernacki et
al. \cite{biernacki2000assessing}. Authors in \cite{Come_Latouche15} proposed
an alternative approach: they introduced an exact version of the ICL for the
stochastic block model, based on a Bayesian approach and maximized it
\emph{directly} with respect to the  number of clusters and to cluster
memberships.     
{ They ran several experiments on simulated and real data showing that maximizing the exact ICL through a greedy search algorithm provided more accurate estimates than those obtained by variational inference or MCMC techniques. Similar results are provided in \cite{wyse2014inferring}, in the context of the latent block model (LBM) for bipartite graphs: the greedy ICL approach outperforms its competitors in both computational terms and in the accuracy of the provided estimates. Therefore, in this paper, we chose to extend the proposed greedy search algorithm to the temporal model.} More    details     are    provided     in    Section
\ref{sec:intro}.

In the following, the expressions ``ICL'' or ``exact
ICL'' will be used interchangeably.  

Following the Bayesian approach, we introduce a prior distribution over the
model parameters $\Phi$ and $\Lambda$, given the meta parameters $K$ and $D$,
denoted $\prior(\Phi,\Lambda|K,D)$. Then the ICL  is the \emph{complete data} log-likelihood given by
\begin{equation}
  \label{eq:ICL}
 ICL(\mathbf{c}, \mathbf{y}, K, D)=\log p(N^{\Delta},\mathbf{c}, \mathbf{y}| K,D),
\end{equation}
where the model parameters $\Phi$ and $\Lambda$ have been integrated out, that is
\begin{equation}
 ICL(\mathbf{c}, \mathbf{y}, K, D)= \log\left( \int p(N^{\Delta},\mathbf{c}, \mathbf{y}| \Lambda, \Phi, K,D)\prior(\Phi,\Lambda|K,D)d\Lambda d\Phi \right).
\end{equation}
We  emphasize  that  the  marginalization over  all  model  parameters
naturally induces  a penalization on  the number of clusters.  For more
details, we refer to \cite{biernacki2000assessing,Come_Latouche15}. The integral can be simplified by a natural independence assumption on the
prior distribution
\begin{equation*}
   \prior(\Lambda, \boldsymbol{\omega}, \boldsymbol{\beta}| K,D)=\prior(\Lambda|K,D) \prior(\boldsymbol{\omega}|K) \prior(\boldsymbol{\beta}|D),
\end{equation*}
which gives
\begin{align}
\label{eq: ICLNI}
 ICL(\mathbf{c}, \mathbf{y}, K, D)=& \log\left(\int p(N^{\Delta}|\Lambda, \mathbf{c}, \mathbf{y},K,D)\prior(\Lambda|K,D)d\Lambda\right) \nonumber \\
 &{}+\log\left(\int p(\mathbf{c}, \mathbf{y}|{\Phi},K,D)\prior({\Phi}|K,D)d{\Phi}\right) \nonumber  \\
 =& \log\left(p(N^{\Delta}| \mathbf{c}, \mathbf{y},K,D)\right) + \log\left( p(\mathbf{c}, \mathbf{y}| K,D)\right). 
\end{align}
Notice that we use in this derivation the implicit hypothesis from equation
\eqref{eq:term_1_ns} which says that $(\mathbf{c}, \mathbf{y})$ is independent
from $\Lambda$ (given $\Phi$, $K$ and $D$). 

\subsection{Conjugated a priori distributions}
A sensible choice of prior distributions over the model parameters is a
necessary condition to have an explicit form of the ICL. 

\subsubsection{Gamma a priori}

In order to integrate out $\Lambda$ and obtain a closed formula for the first term on the right hand side of \eqref{eq: ICLNI}, we impose a Gamma a priori distribution over $\Lambda$
\begin{equation*}
  \prior(\lambda_{kgd}| a, b)=\frac{{b}^{a}}{\Gamma(a)}\lambda_{kgd}^{a-1}e^{-b\lambda_{kgd}},
\end{equation*}
leading to following joint density
\begin{equation}
\label{eq:gamma_ap}
  \prior(\Lambda|K,D)=\prod_{k,g,d} p(\lambda_{kgd}| a, b),
\end{equation}
where $a,b>0$ and $\Gamma(\bullet)$ is the gamma function. By multiplying \eqref{eq:obsLL} and \eqref{eq:gamma_ap}, the
joint density for the pair $(N^{\Delta},\Lambda)$ follows
\begin{equation*}
  p(N^{\Delta},\Lambda | \mathbf{c}, \mathbf{y},K,D) =
\prod_{k,g,d}\left[ \frac{b^a}{\Gamma(a)P_{kgd} }e^{- \lambda_{kgd} [R_{kgd} + b]}\lambda_{kgd}^{S_{kgd}+a -1}\right].
\end{equation*}
This quantity can now be easily integrated w.r.t. $\Lambda$ to obtain
\begin{equation}
 \label{eq:ICLI1}
   p(N^{\Delta}| \mathbf{c}, \mathbf{y},K,D) =  \prod_{k,g,d}L_{kgd},
 \end{equation}
with
 \begin{equation}
   \label{eq:ICLLkgd}
L_{kgd}=\frac{b^a}{\Gamma(a) P_{kgd} }\frac{\Gamma(S_{kgd}+a)}{[R_{kgd}+ b]^{S_{kgd} + a}}.
 \end{equation}
 A non informative prior for the Poisson distribution corresponds to limiting
 cases of the Gamma family, when $b$ tends to zero. In all the experiments we
 carried out, we set the parameters $a$ and $b$ to one, in order to have
 unitary mean and variance for the Gamma distribution.

\subsubsection{Dirichlet a priori}

We attach a factorizing Dirichlet a priori distribution to $\Phi$, namely
\begin{align*}
 \prior(\Phi|K,D) =& \text{Dir}_K(\boldsymbol{\omega}; \alpha,\dots,\alpha) \times \text{Dir}_D(\boldsymbol{\beta}; \gamma,\dots,\gamma),
\end{align*}
where the parameters of each distribution have been set constant for
simplicity. It can be proved (\ref{sec:joint-integr-dens}) that the joint
integrated density for the pair $(\mathbf{c},\mathbf{y})$, reduces to 
\begin{equation}
\label{eq:Dir}
p(\mathbf{c}, \mathbf{y}| K,D)=\frac{\Gamma(\alpha K)}{\Gamma(\alpha)^K}\frac{\prod_{k\leq K}\Gamma(|A_k| + \alpha)}{ \Gamma(N + \alpha K)}
\frac{\Gamma(\gamma D)}{\Gamma(\gamma)^D}\frac{\prod_{d\leq D}\Gamma(|C_d| + \gamma)}{ \Gamma(U + \gamma D)}.
\end{equation}
A common choice consists in fixing these parameters to 1 to get a uniform
distribution, or to $1/2$ to obtain a Jeffreys non informative prior.

\section{ICL maximization}
\label{sec:greedy_search}
The integrated complete likelihood (ICL) in equation \eqref{eq: ICLNI} has to
be maximized with respect to the four unknowns $\mathbf{c}$, $\mathbf{y}$, $K$,
and $D$ which are discrete variables. Obviously no closed formulas can be obtained
and it would computationally prohibitive to test every combination of the four
unknowns. Following the approach described in \cite{Come_Latouche15}, we rely on a greedy search strategy. The main idea is to start with a
fine clustering of the nodes and of the intervals (possibly size one clusters)
and then to alternate between an exchange phase where nodes/intervals can move
from one cluster to another and a merge phase where clusters are
merged. Exchange and merge operations are locally optimal and are guaranteed
to improve the ICL. 

The algorithm is described in detail in the rest of the section. An analysis
of its computational complexity is provided in \ref{sec:comp-compl}. 

\begin{remark}
  The algorithm is guaranteed to increase the ICL at each step and thus to
  converge to a local maximum. Randomization can be used to explore several
  local maxima but the convergence to a global maximum is not guaranteed.
{ Moreover, let us denote by $\hat{\mathbf{c}}$, $\hat{\mathbf{y}}$, $\hat{K}$, $\hat{D}$ the estimators of $\mathbf{c}$, $\mathbf{y}$, $K$
and $D$, respectively, obtained through the maximization of the function in equation \eqref{eq: ICLNI}. A formal proof of the consistency of these estimators is outside the scope of this paper. More in general, the consistency of this kind of estimators, maximizing the exact ICL, is still an open issue. 
}
\end{remark}
\subsection{Initialization} 
Initial values are fixed for both $K$ and $D$, say $K_{\max}$ and
$D_{\max}$. These values may be fixed equal to $N$ and $U$ respectively and
each node (interval) would be alone in its own cluster (time
cluster). Alternatively, simple clustering algorithms (k-means, hierarchical
clustering) may be used to reduce $K_{\max}$ and $D_{\max}$ up to a certain
threshold. This choice should be preferred to speed up the greedy search.

\subsection{Greedy - Exchange (GE)} 
A shuffled sequence of all the nodes (time intervals) in the graph is
created. One node (time interval) is chosen and is moved from its current
(time) cluster into the (time) cluster leading to the highest increase in the
exact ICL, if any. This is called a greedy exchange (GE). This routine is
applied to every node (time interval) in the shuffled sequence. This iterative
procedure is repeated until no further improvement in the exact ICL is
possible.
 In case a node (time interval) is alone inside its cluster, an exchange becomes a merge of two clusters (see below).

 The ICL does not have to be completely evaluated before and after each swap:
 possible increases can be computed directly, reducing the computational
 cost. Let us consider first the case of temporal intervals. Moving interval $I_u$ from the cluster
 $C_{d'}$ to cluster $C_{l}$ induces a modification of the ICL given by
\begin{align*}
 \Delta^{E,T}_{d'\rightarrow l}:=&  ICL(\mathbf{c}, \mathbf{y}^*, K,D)-  ICL(\mathbf{c}, \mathbf{y},K,D), \\
 =&  \left[\log\left( \prior(\mathbf{c}, \mathbf{y}^*| K,D)\right) + \sum_{k,g,d}\log( L_{kgd}^*) \right] \\
 & - \left[\log\left( \prior(\mathbf{c}, \mathbf{y}| K,D)\right) + \sum_{k,g,d}\log( L_{kgd}) \right],
 \end{align*}
where  $y^*$ and $L_{kgd}^*$ refer to the new configuration where $I_u \in
C_l$. It can easily be shown that $\Delta_{d'\rightarrow l}$ reduces to 
\begin{multline}
 \Delta^{E,T}_{d'\rightarrow l}=\\
 \label{eq:exchange}
\log\left(\frac{\Gamma(|C_{d'}| - 1 + \gamma)\Gamma(|C_l|+1 +\gamma)}{\Gamma(|C_{d'}| + \gamma)\Gamma(|C_l|+\gamma)}\right)+\sum_{k,g}\log\left(\frac{L_{kgd'}^{*}L_{kgl}^{*}}{L_{kgd'}L_{kgl}}\right).
\end{multline}
The case of nodes is slightly more complex. When a node is moved from
cluster $A_{k'}$ to $A_l$, with $k' \neq l$, the change in the ICL is
 \begin{equation*}
  \Delta^{E,V}_{k' \rightarrow l} := ICL(\mathbf{c^*}, \mathbf{y}, K,D)- ICL(\mathbf{c},\mathbf{y},K,D),
 \end{equation*}
which simplifies into
 \begin{align*}
 \Delta^{E,V}_{k'\rightarrow l} =& \log\left(\frac{\Gamma(|A_{k'}| - 1 + \alpha)\Gamma(|A_l|+1 +\alpha)}{\Gamma(|A_{k'}| + \alpha)\Gamma(|A_l|+\alpha)}\right) \\
   &+\sum_{g \leq K}\sum_{d \leq D}\log(L^*_{k'gd}) + \sum_{g \leq K}\sum_{d \leq D} \log(L^*_{lgd}) \\
   &+\sum_{k \leq K}\sum_{d \leq D}\log(L^*_{kk'd})+\sum_{k \leq K}\sum_{d \leq D}\log(L^*_{kld})\\
   &-\sum_d (\log(L^*_{k'k'd}) + \log(L^*_{k'ld}) +\log(L^*_{lk'd}) + \log(L^*_{lld})) \\
   &-\sum_{g \leq K}\sum_{d \leq D}\log(L_{k'gd}) - \sum_{g \leq K}\sum_{d \leq D} \log(L_{lgd}) \\
   &-\sum_{k \leq K}\sum_{d \leq D}\log(L_{kk'd})- \sum_{k \leq K}\sum_{d \leq D}\log(L_{kld})\\
   &+\sum_d (\log(L_{k'k'd}) + \log(L_{k'ld}) +\log(L_{lk'd}) + \log(L_{lld})),
 \end{align*}
where $\mathbf{c^*}$ and $L_{kgd}^*$ refer to the new configuration.

\subsection{Greedy - Merge (GM)} 
Once the GE step is concluded, all possible merges of pairs of clusters (time
clusters) are tested and the best merge is finally retained. This is called a
greedy merge (GM).This procedure is repeated until no further improvement in
the ICL is possible.

In this case too, the ICL does not need to be explicitly computed. Merging in fact
time clusters $C_{d'}$ and $C_l$ into $C_l$ leads to the following ICL modification
\begin{align}
\label{eq:merge}
\Delta^{M,T}_{d' \rightarrow l}:=&  ICL(\mathbf{c},\mathbf{y}^*,K,D-1)-  ICL(\mathbf{c}, \mathbf{y},K,D) \nonumber \\
                                          =& \log\left(\frac{\prior(\mathbf{c}, \mathbf{y}^*| K,D-1)}{\prior(\mathbf{c}, \mathbf{y}| K,D)} \right)+ \sum_{k,g}\left((\log(L_{kgl}^*)-\log(L_{kgd'}L_{kgl})\right)
\end{align}
Notice that if $d\leq l$, then $l$ has to be replaced by $l-1$ inside $L_{kgl}^*$. 

 When merging clusters $A_{k'}$ and $A_l$ into the cluster $A_l$, the change in the ICL can be expressed as follows
\begin{align*}
 \Delta^{M,V}_{k'\rightarrow l} :=& ICL(\mathbf{c}^*, \mathbf{y}, K-1,D)- ICL(\mathbf{c}, \mathbf{y},K,D) =\\
 =&\log\left(\frac{\prior(\mathbf{c^*}, \mathbf{y}| K-1,D)}{\prior(\mathbf{c}, \mathbf{y}| K,D)} \right)+ \\
  & +\sum_{g \leq K}\sum_{d \leq D}(\log(L^*_{lgd}) + \log(L^*_{kld})) - \sum_d \log(L^*_{lld})\\
  &-\sum_{g \leq K}\sum_{d \leq D}\log(L_{k'gd}) - \sum_{g \leq K}\sum_{d \leq D} \log(L_{lgd}) \\
  &-\sum_{k \leq K}\sum_{d \leq D}\log(L_{kk'd})- \sum_{k \leq K}\sum_{d \leq D}\log(L_{kld})\\
  &+\sum_d (\log(L_{k'k'd}) + \log(L_{k'ld}) +\log(L_{lk'd}) + \log(L_{lld})).
 \end{align*}

\subsection{Optimization strategies}\label{sec:optim-strat}
We have to deal with two different issues:
\begin{enumerate}
 \item  the optimization order of nodes and times: we could either run the
   greedy algorithm for nodes and times separately or choose an hybrid strategy
   that switches and merges nodes and time intervals alternatively, for instance;
 \item  whether to execute merge or switching movements at first.
\end{enumerate}
The second topic has been largely discussed in the context of modularity
maximization for community detection in static graphs. One of the most
commonly used algorithms is the so-called Louvain method
\citep{Blondel08fastunfolding} which proceeds in a rather similar way as the
one chosen here: switching nodes from clusters to clusters and then merging
clusters. This is also the strategy used in \citep{Come_Latouche15} for
stationary SBM. Combined with a choice of sufficiently small values of  $K_{max}$
and $D_{max}$, this approach gives very good results at a reasonable
computational cost. It should be noted that more complex approaches based on
multilevel refinements of a greedy merge procedure have been shown to give
better results than the Louvain method in the case of modularity maximization
(see \citep{NoakRotta}). However, the computation complexity of those
approaches is acceptable only because of the very specific nature of the
modularity criterion and with the help of specialized data structures. We
cannot leverage such tools for ICL maximization. 

The first issue is hard to manage since the shape of the function
$ICL(\mathbf{c},\mathbf{y}, K, D)$ is unknown. We developed three optimization strategies:
\begin{enumerate}
 \item \emph{GE} + \emph{GM} for time intervals and then \emph{GE} + \emph{GM} for nodes (\textbf{Strategy A});
 \item \emph{GE} + \emph{GM} for nodes and then \emph{GE} + \emph{GM} for times (\textbf{Strategy B});
 \item Mixed \emph{GE} + mixed \emph{GM} (\textbf{Strategy C}).
\end{enumerate}     
In the mixed \emph{GE} a node is chosen in the shuffled sequence of nodes and
moved to the cluster leading to the highest increase in the ICL. Then a time
interval is chosen in the shuffled sequence of time intervals and placed in
the best time cluster and so on alternating between nodes and time intervals
until no further increase in the ICL is possible. The mixed \emph{GM} works
similarly. In all the experiments, the three optimization strategies are
tested and the one leading to the highest ICL is retained.

\section{Experiments}\label{sec:experiments}
To assess the reliability of the proposed methodology some experiments on synthetic and real data were conducted. All runtimes mentioned in the next two sections are measured on a twelve cores Intel Xeon server with 92 GB of
main memory running a GNU Linux operating system. The greedy algorithm described in Section \ref{sec:greedy_search} was implemented in C++. An euclidean hierarchical clustering algorithm was used to initialize the labels and $K_{max}$ and $D_{max}$ have been set equal to $N/2$ and $U/2$ respectively. 

\subsection{Simulated Data}
\subsubsection{First Scenario} 
We simulated interactions between 50 nodes, belonging to three clusters
$A_1, A_2, A_3$. Interactions take place over 50 times intervals of unitary
length, belonging to three time clusters (denoted $C_1, C_2, C_3$).
Clusters are assumed to be balanced on average by fixing
$\boldsymbol{\omega}=\boldsymbol{\beta}=(\frac{1}{3},\frac{1}{3},\frac{1}{3})$. Notice
that while the clusters are balanced on \emph{average} they can be relatively
imbalanced in some particular cases. 

A community structure setting is chosen, corresponding to the following
diagonal form for the intensity matrix $L$
\begin{equation*}
     L=
  \begin{pmatrix}
     \psi &  2 &  2  \\
      2 & \psi & 2 \\
      2 & 2 & \psi  
  \end{pmatrix},
\end{equation*}
where $\psi$ is a free parameter in $[2, +\infty)$. A non stationary behaviour
is obtained by modifying the intensity matrix over time as follows
\begin{equation}
\label{eq:intensity_matrix}
 \Lambda(u)=L\mathbf{1}_{C_1}(u) + \sqrt{\gamma} L \mathbf{1}_{C_2}(u) + \gamma L \mathbf{1}_{C_3}(u), \quad u \in \{1, \dots, 50\} 
\end{equation}
where $\gamma$ is a free parameter in $[1, \infty)$ and $\mathbf{1}_A$ denotes
the indicator function over a set $A$. In other words, $\Lambda(u)$ is equal
to $L$ when $u$ belongs to $C_1$, to $\sqrt{\gamma} L$ when $u$ belongs to
$C_2$ and to $\gamma L$ when $u$ belongs to $C_3$. The overall community
pattern does not evolve through time but the average interaction intensity is
different in the three time clusters. Both the community structure and the non
stationary behaviour can be made more or less obvious based on the value of
$\psi$ and $\gamma$. 

For several values of the
pair ($\psi,\gamma$), 50 dynamic graphs were sampled according to the Poisson
intensities in equation \eqref{eq:intensity_matrix}. Estimates of labels
vectors $\mathbf{y}$ and $\mathbf{c}$ are provided for each graph\footnote{The
  average runtime of our implementation on those artificial data is
  \emph{0.96 second}.}. The greedy algorithm following the
optimization strategy \textbf{A}, led to the best results (see next paragraph for more
details). In order to avoid convergence to local maxima, ten estimates of
labels are provided for each graph and the pair
($\hat{\boldsymbol{y}}, \hat{\boldsymbol{c}}$) leading to the highest ICL is
retained\footnote{Calculations are done in parallel as they are
  independent. The reported runtime is the wall clock time.}.

\begin{figure*}[ht]
\centering
\begin{subfigure}{.5\textwidth}
  \centering
  \includegraphics[width=\linewidth]{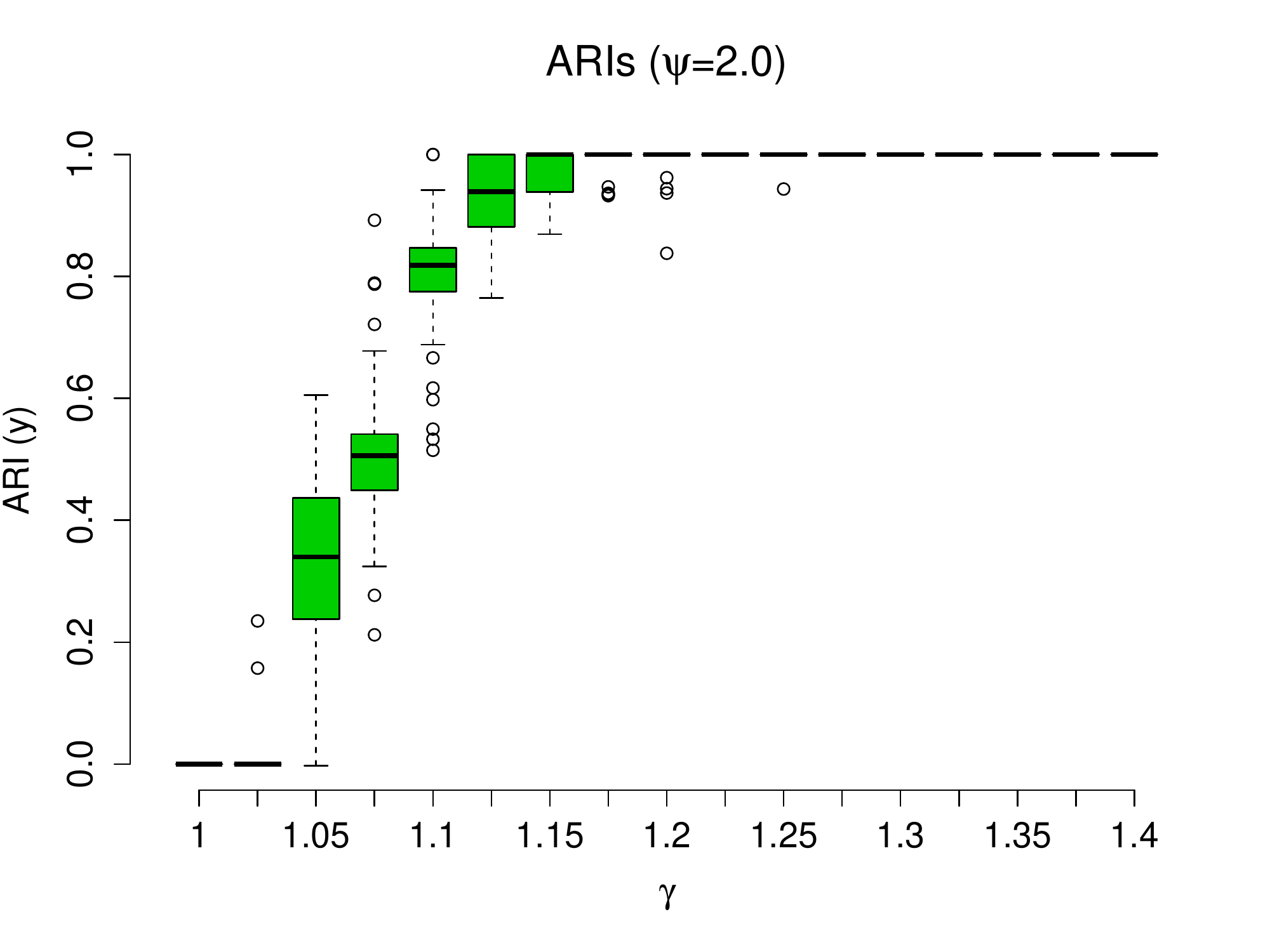}
  \captionsetup{format=hang}
  \subcaption{\footnotesize Time clustering}
  \label{fig:Ari_y}
\end{subfigure}%
\begin{subfigure}{.5\textwidth}
  \centering
  \includegraphics[width=\linewidth]{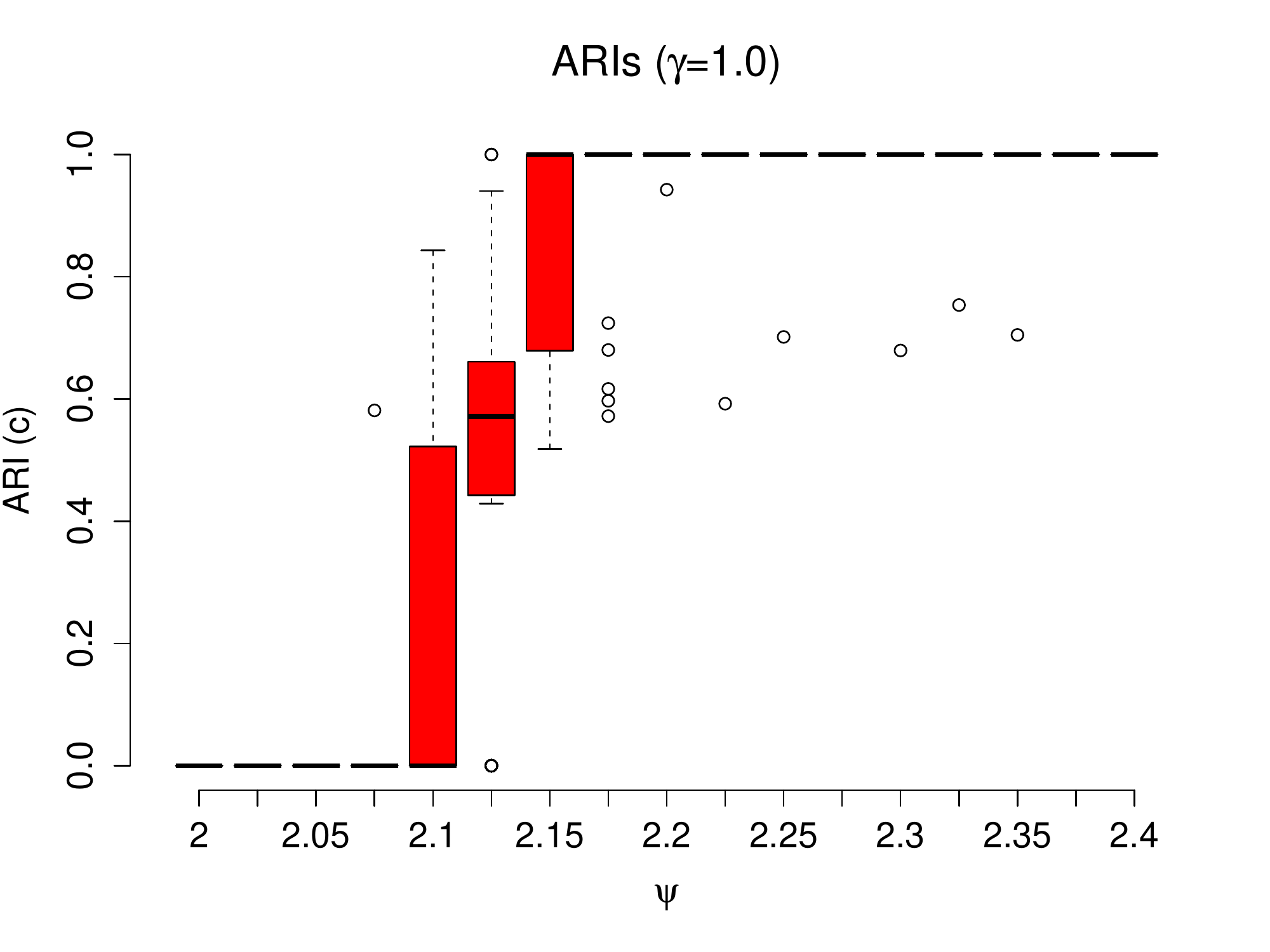}
  \captionsetup{format=hang}
  \subcaption{\footnotesize Nodes clustering}
  \label{fig:Ari_c}
\end{subfigure}
\caption{Box plots of ARIs for both clusterings of nodes and time intervals. Both clusterings reach the maximum effectiveness for higher values of contrast parameters.}
\label{fig:ari_s}
\end{figure*}

Experiments show that for sufficiently large values of $\psi$ and $\gamma$,
the true structure can always be recovered. We can see this in detail for two
special cases, as illustrated in Figure \ref{fig:ari_s} .

In Figure \ref{fig:Ari_y}, we set $\psi=2$, which means there is not any
community structure and let $\gamma$ varying in the range
$[1, 1.05, \dots, 1.4]$. Adjusted Rand Indexes (ARIs) \cite{rand1971objective}
are used to assess the time clustering, varying between zero (null clustering)
and one (optimal clustering). When $\gamma=1$ we are in a degenerate case and
no time structure affects the interactions: not surprisingly the algorithm
assigns all the intervals to the same cluster (null ARI). The higher the value
of $\gamma$ the more effective the clustering is up to a perfect recovery of
the planted structure (ARI of 1). In particular the true time structure is
fully recovered for all the fifty graphs when $\gamma$ is higher than $1.3$.

\begin{figure}[ht]
 \centering
 \includegraphics[width=\linewidth]{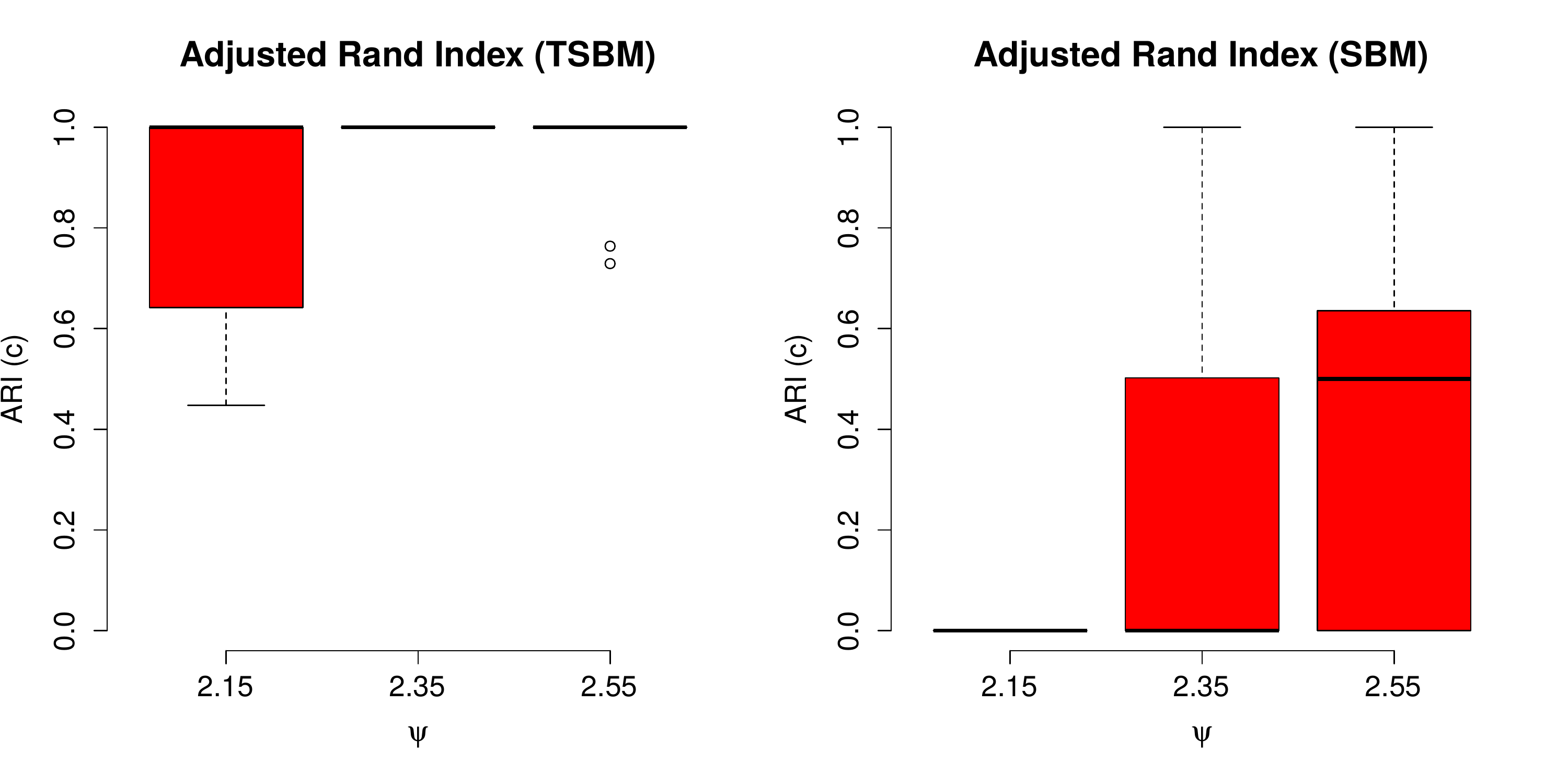}
 \caption{Comparison between the temporal SBM we propose and a classical SBM in a stationary context (any time cluster).}
\label{fig:TSBMvsSBM}
\end{figure}

Similar results can be observed in Figure \ref{fig:Ari_c} about nodes
clustering: by setting $\gamma=1$, we removed any time structure and a
stationary community structure is detected by the model. In this case it is
interesting to make a comparison with a traditional SBM, which is expected to
give similar results to those shown in Figure \ref{fig:Ari_c}. For a fixed
value of $\psi$ we simulated a dynamic graph, corresponding to 50 adjacency
matrices, one per time interval. Then a static graph is obtained by summing up
these adjacency matrices. The temporal SBM (TSBM) we propose deals with the
dynamic graph, whereas a SBM is used on the static graph \footnote{This choice is the most natural one to compare the two models. Alternatively, the SBM could be used on a single adjacency matrix among the fifty adjacency matrices provided, at each iteration. In the experiments we carried out, we obtained similar results for the two options.}. The
Gibbs sampling algorithm introduced in \cite{nouedoui2013} was
used to recover the number of clusters and cluster memberships according to a
SBM (with Poisson distributed edge values). The experiment was repeated 50
times for each value of $\psi$ in the set $\{2.15, 2.35, 2.55\}$. In Figure
\ref{fig:TSBMvsSBM} we compare the ARIs of the two models for each value of
$\psi$. 

The greedy ICL TSBM (faster than
the Gibbs sampling algorithm, who has an average runtime of \emph{15.15 seconds}) recovers the true structure at levels of contrast lower
than those required by the Gibbs sampling algorithm (SBM). {
This comparison aims at showing that, in a stationary framework, the TSBM works at least as well as a standard SBM. The difference in terms of performance of the two models in this context can certainly be explained by the greedy search approach which is more effective than Gibbs sampling, as expected (see \citep{Come_Latouche15} and section \ref{sec:intro}).}

\subsubsection{Optimization strategies} 
\begin{figure}[ht]
 \centering
 \includegraphics[width=\linewidth]{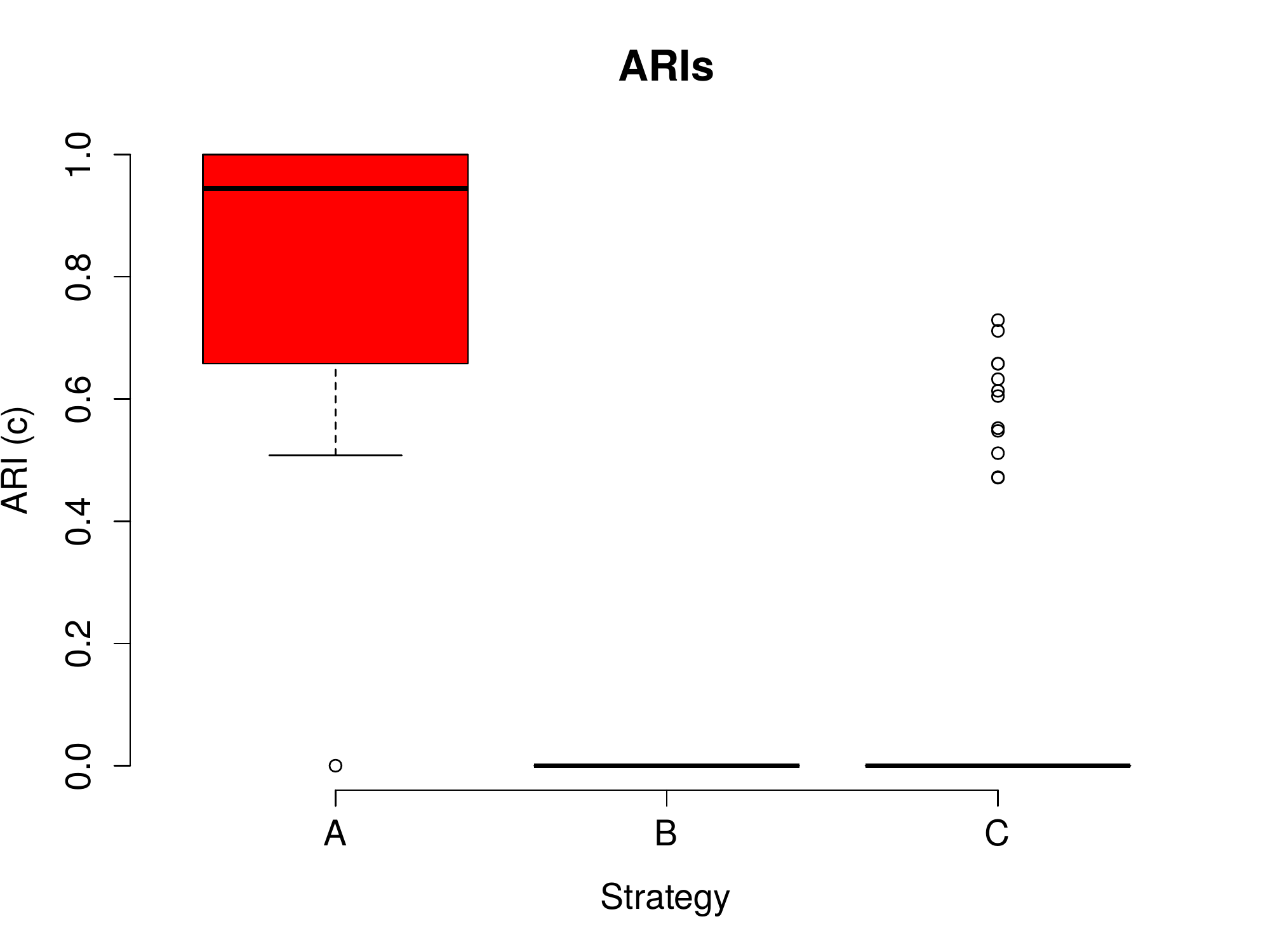}
 \caption{Box plots of 50 ARIs for clustering of nodes for each optimization strategy in the first scenario with $\psi=2.15$.}

\label{fig:comp_paths}
\end{figure}

As mentioned in the previous section, in the present experiments, the optimization
strategy \textbf{A} is more efficient than the two other strategies outlined
in Section \ref{sec:optim-strat}. We illustrate this superiority in the
following test: the pair $(\gamma, \psi)$ is set to $(1, 2.15)$ and 50 dynamic
graphs are simulated according to the same settings discussed so far. Three
different estimations are obtained, one for each strategy, and ARIs for nodes
labels are computed. Results in Figure (\ref{fig:comp_paths}) can be compared with
the mean value of the final ICL for each strategy:

\medskip
\begin{center}
 \begin{tabular}{c|c}
  \hline
  		 & \textbf{mean ICL} \\
  \hline
   \footnotesize\textbf{strategy A} & $-70845.64$ \\
  \hline  
   \footnotesize\textbf{strategy B} & $-70894.67$ \\
  \hline  
   \footnotesize\textbf{strategy C} & $-70885.22$ \\		
  \hline 
 \end{tabular}
\end{center}
\medskip

\subsubsection{Scalability}
A full scalability analysis of the proposed algorithm is out of the
scope of this paper (see \ref{sec:comp-compl}), but we have performed a limited
assessment in this direction with a simple example. 

\begin{figure}[ht]
 \centering
 \includegraphics[width=\linewidth]{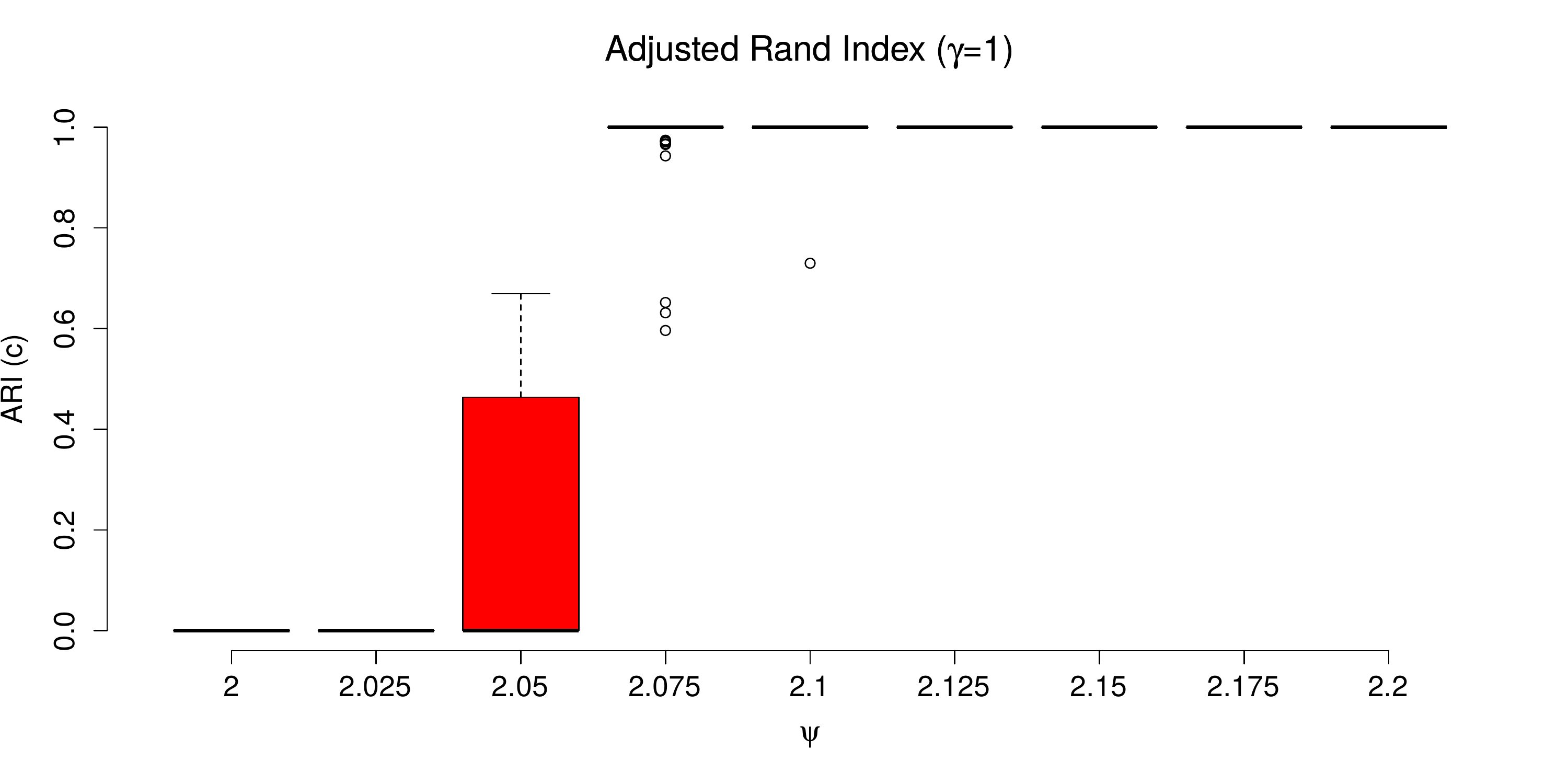}
 \caption{Box plots of 50 ARIs for clustering of nodes in the first scenario, with $N=100$ and $U=100$.}
\label{fig:ARIc_higher_NU}
\end{figure}

A fixed $\gamma=1$ is maintained and for several values of $\psi$ and $50$ dynamic
graphs with 100 nodes and 100 times intervals were sampled according to the
intensity in equation \eqref{eq:intensity_matrix}. The mean runtime for
reading and providing labels estimates for $each$ dynamic graph is \emph{13.16
seconds}. As expected, the algorithm needs a lower contrast to recover the true structure as
the reader can observe by comparing Figure (\ref{fig:ARIc_higher_NU}) with
Figure (\ref{fig:Ari_c}). This is a consequence of the increase in the number
of interactions (induced by the longer time frame). 

In terms of computational burden, each dynamic graph is handled in a average
time of \emph{13.16 seconds}, that is less than 14 slower than in the case of a graph
with 50 nodes and 50 time intervals. As we use $K_{max}=N/2$ and $D_{max}=U/2$,
the worst case cost of one ``iteration'' of the algorithm is $O((N+U)UN^2)$
and thus doubling both $N$ and $U$ should multiply the runtime by 16. On this
limited example, the growth is slightly less than expected. 

\subsubsection{Non community structure} 
\begin{figure}[ht]
 \centering
 \includegraphics[width=\linewidth]{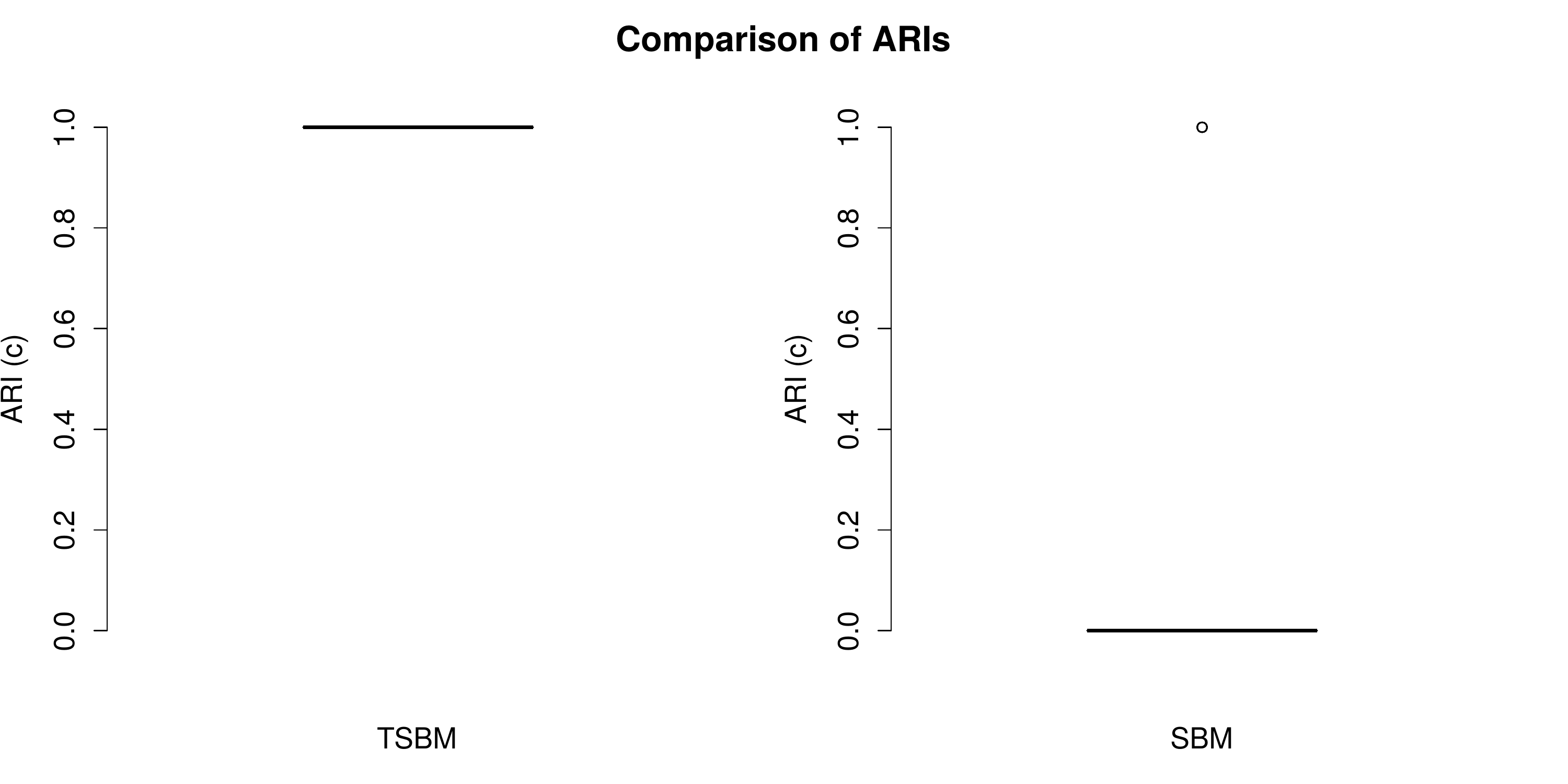}
 \caption{Comparison between the temporal SBM and a SBM in the second scenario.}
\label{fig:TSBMvsSBM_sc2}
\end{figure}

We now consider a different scenario showing how the TSBM model can perfectly
recover a clustering structure in a situation where the SBM
fails. We considered two clusters of nodes $A_1$ and $A_2$ and two time
clusters $C_1$ and $C_2$ (clusters are balanced in average as in the previous
examples). We simulated directed interactions between $50$ nodes over $100$
time intervals according to the following intensity matrix
\begin{equation} 
\Lambda(u)=L_1 \mathbf{1}_{C_1}(u) + L_2 \mathbf{1}_{C_2}(u), \qquad u \in \{1,\dots, 100\},
\label{eq:intensity_matrix_sc2}
\end{equation}
where
\begin{equation*}
     L_1=
  \begin{pmatrix}
     2 &  1  \\
     1 &  2  \\
  \end{pmatrix}
\qquad\text{and}\qquad
     L_2=
  \begin{pmatrix}
     1 &  2  \\
     2 &  1  \\
  \end{pmatrix}.
\end{equation*}
In this scenario, a clustering structure is persistent over time, but the
agents behaviour changes abruptly depending on the time cluster the
interactions are taking place, moving from a community like pattern to a
bipartite like one.  When aggregating observations, since the
expected percentage of time intervals belonging to cluster $C_1$ is $50\%$,
the two opposite effects compensate each other (on average) and the SBM cannot
detect any community structure. This can be seen in Figure
\ref{fig:TSBMvsSBM_sc2}: we simulated 50 dynamic graphs according to the
Poisson intensities in equation \eqref{eq:intensity_matrix_sc2} and estimates
of $\mathbf{c}$ and $K$ are provided for each graph by both TSBM and SBM. The
outliers ARIs in the right hand side figure (7 over 50) correspond to sampled
vectors $\mathbf{y}$ in which the proportion of time intervals belonging to
cluster $C_1$ is far from $1/2$. No outlier is observed when 
the experiment is performed with a \emph{fixed} label vector $\mathbf{y}$ placing the same
number of time intervals in each cluster.

The optimization strategy \textbf{A} has been used to produce the results
shown in Figure \ref{fig:comp_paths}. Very similar results can be obtained
through optimization strategies \textbf{B} and \textbf{C}: with these settings the
greedy ICL algorithm can always estimate the true vectors $\mathbf{c}$ and
$\mathbf{y}$.

\begin{figure*}[ht]
\centering
\begin{subfigure}{.5\textwidth}
  \centering
  \includegraphics[width=\linewidth]{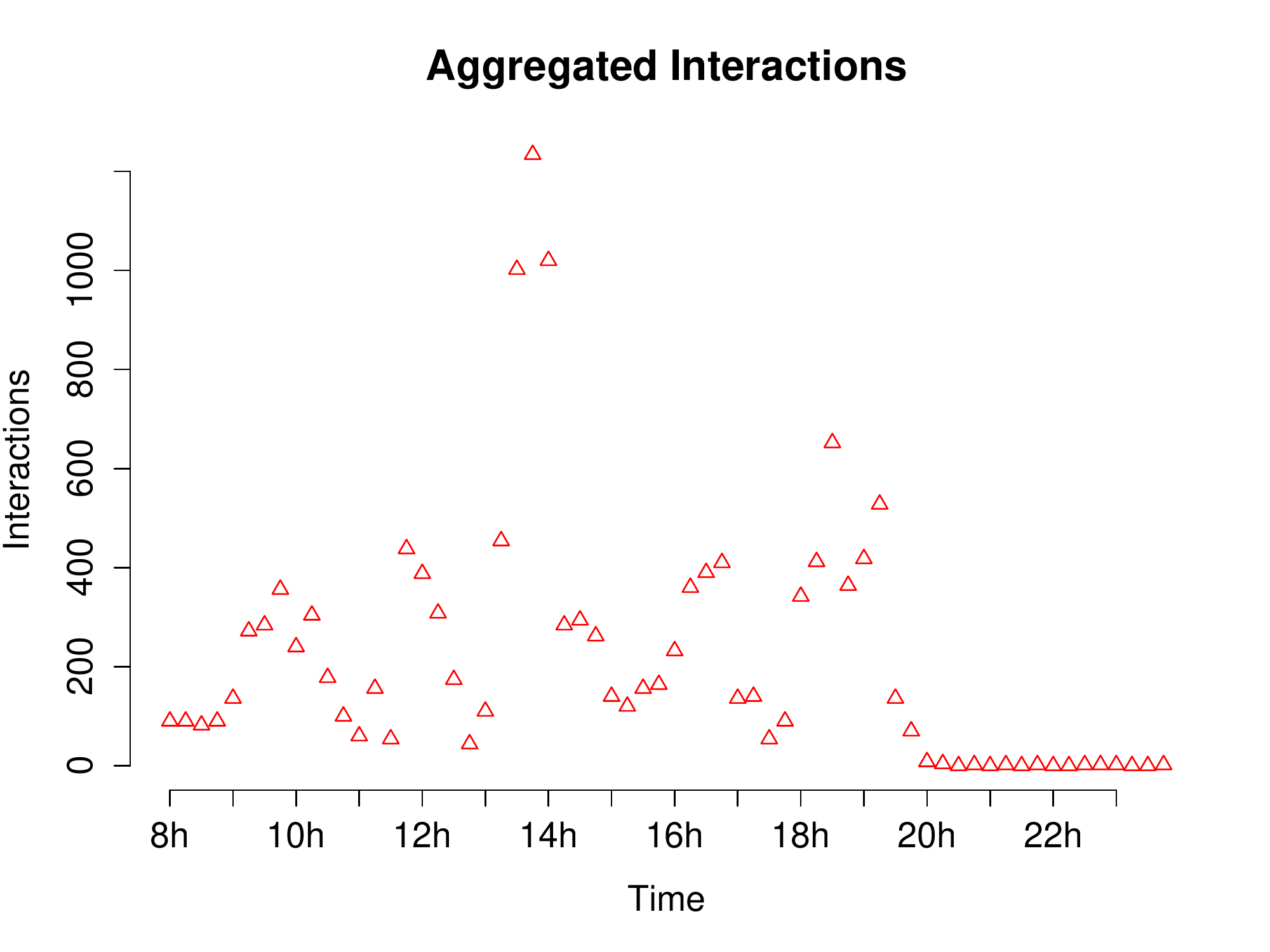} 
  \caption{\footnotesize Aggregated connections.}
  \label{fig:Agginter}
\end{subfigure}%
\begin{subfigure}{.5\textwidth}
  \centering
  \includegraphics[width=\linewidth]{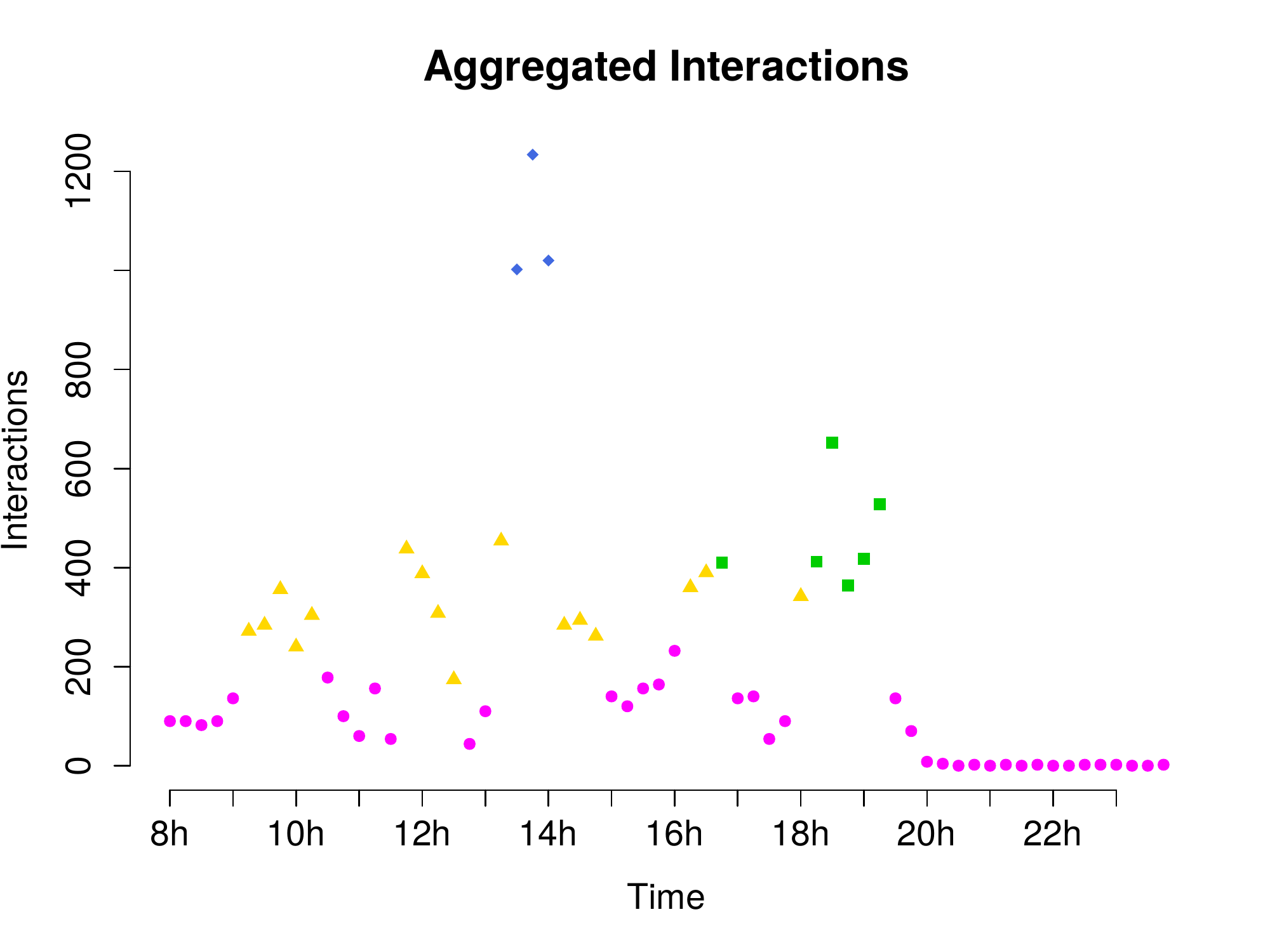}
  \caption{\footnotesize Clustered time intervals.}
  \label{fig:TC}
\end{subfigure}
\caption{\small Aggregated connections for each time interval (\ref{fig:Agginter}) and time clusters found by our model (\ref{fig:TC}) are compared.}
\label{fig:MainFIgure}
\end{figure*}

\subsection{Real Data}

The data set we used was collected during the \textbf{ACM Hypertext} conference held in Turin, June 29th - July 1st, 2009. It represents the dynamic network of face-to-face proximity interactions of 113 conference attendees over about 2.5 days\footnote{More informations can be found at: 

\url{http://www.sociopatterns.org/data sets/hypertext-2009-dynamic-contact-network/ }.

}


We focused on the first conference day, namely the twenty four hours going
from 8am of June 29th to 7.59am of June 30th. The day was partitioned in small
time intervals of 20 seconds in the original data frame and interactions of
face-to-face proximity (less than 1.5 meters) were monitored by electronic
badges that attendees volunteered to wear. Further details can be found in
\cite{Isella:2011qo}.  We considered 15 minutes time aggregations, thus
leading to a partition of the day made of 96 consecutive quarter-hours ($U=96$
with previous notation). A typical row of the aggregated data set looks like
the following one:

\medskip
\begin{center}
 \begin{tabular}{c|c|c|c}
  \hline
  \footnotesize\emph{ID$1$} & \footnotesize\emph{ID$2$} & \footnotesize\emph{Time Interval ($15m$)} & \footnotesize\emph{Number of interactions} \\
  \hline
   52 & 26 & 5 & 16 \\
  \hline 
 \end{tabular}
\end{center}
\medskip

It means that conference attendees 52 and 26, between 9am and 9.15am, have spoken for $16 \times 20s \approx 5m30s$.  

In Figure \ref{fig:Agginter}, we computed the total number of interactions for each quarter hour. The presence of a time pattern is clear: the volume of interactions, for example, is much higher at $14$pm than at $9$am.
The greedy ICL algorithm found 20 clusters for nodes (people) and 4 time clusters. 
Figure \ref{fig:TC} shows how daily quarter-hours are assigned to each
cluster: it can clearly be seen how time intervals corresponding to the
highest number of interactions have been placed in cluster $C_4$, those
corresponding to an intermediate interaction intensity, in $C_2$ (yellow) and
$C_3$ (green). Cluster $C_1$ (magenta) contains intervals marked by a weaker
intensity of interactions. It is interesting to note how the model closely
recovers times of social gathering\footnote{A complete program of the day can be found at \url{http://www.ht2009.org/program.php}.}: 
\begin{itemize}
 \item 9.00-10.30 - set-up time for posters and demos.
 \item 13.00-15.00 - lunch break.
 \item 18.00-19.00 - wine and cheese reception.
\end{itemize}


Results in Figure \ref{fig:MainFIgure} are obtained through the optimization strategy \textbf{A}. 
To make a comparison with the other two optimization strategies, we run the algorithm ten times for each strategy (\textbf{A}, \textbf{B} and \textbf{C}) and compare the final values of the ICL.
Labels $\mathbf{c}$ and $\mathbf{y}$ are randomly initialized before each run, according to multinomial distributions (no hierarchical clustering was used)
and $K_{\max}$ and $D_{\max}$ are set equal to $N/2$ and $U/2$, respectively. 
The mean final values of the ICL are reported in the following table:   

\medskip
\begin{center}
\label{tab:paths_real}
 \begin{tabular}{c|c}
  \hline
  		 & \textbf{mean ICL} \\
  \hline
   \footnotesize\textbf{Strategy A} & $-32746.51$ \\
  \hline  
   \footnotesize\textbf{Strategy B} & $-33072.99$ \\
  \hline  
   \footnotesize\textbf{Strategy C} & $-32116.01 $ \\		
  \hline 
 \end{tabular}
\end{center}
\medskip

\begin{figure}[ht]
 \centering
 \includegraphics[width=\linewidth]{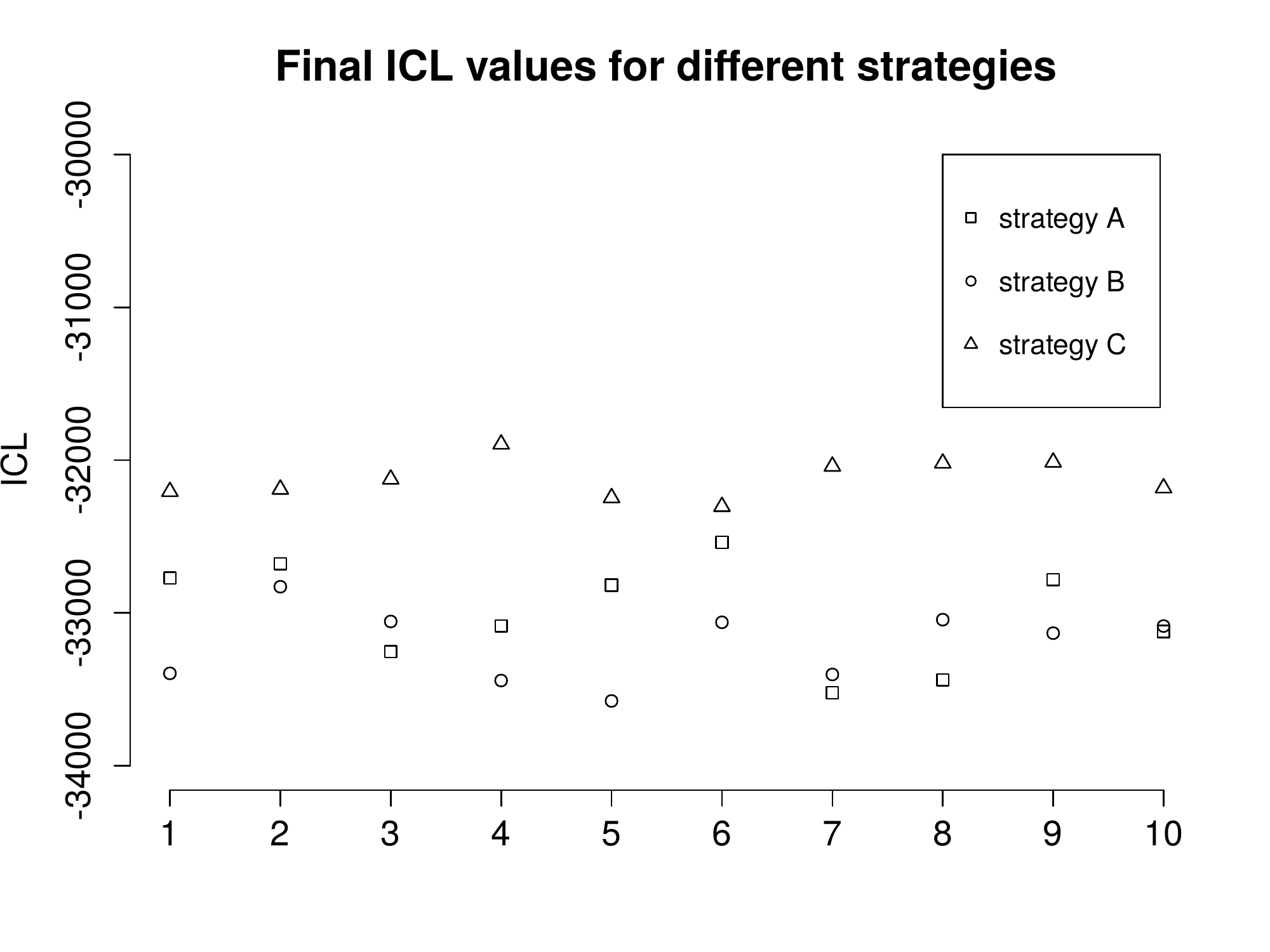}
 \caption{Comparison between the final values of the ICL obtained through different optimization strategies. On the horizontal axis we have the index of the experiment, on the vertical axis the final value of the ICL for each strategy.}
\label{fig:3_paths_rd}
\end{figure}

As it can be seen, the hybrid strategy \textbf{C} is the one leading to the highest final ICL, on average. In Figure (\ref{fig:3_paths_rd}) we report the final value of the ICL for each run (from 1 to 10) for each strategy. The optimization strategy \textbf{C} always outperforms the remaining two patterns. 

\section{Conclusion}
We proposed a non-stationary extension of the stochastic block model (SBM)
allowing us to simultaneously cluster nodes and infer the time structure of a
network. The approach we chose consists in partitioning the time interval over
which interactions are studied into sub-interval of fixed identical
duration. Those intervals provide aggregated interaction counts that are
studied with a SBM inspired model: nodes and time intervals are clustered in
such a way that aggregated interaction counts are homogeneous over clusters.
We derived an exact integrated classification likelihood (ICL) for such a
model and proposed to maximize it with a greedy search strategy. The
experiments we run on artificial and real world networks highlight the
capacity of the model to capture non-stationary structures in dynamic graphs.

\newpage

\appendix\label{appendix}

\section{Joint integrated density for labels}\label{sec:joint-integr-dens}
Consider at first the vector $\mathbf{c}$, whose joint probability function is given by \eqref{eq:clabel}. We attach a Dirichlet a priori distribution to the $K$-vector $\boldsymbol{\omega}$
\begin{equation*}
\prior(\boldsymbol{\omega} | \alpha, K)= \frac{\Gamma(\alpha K)}{\Gamma(\alpha)^K} \prod_{k=1}^K \omega_k^{\alpha - 1}.
\end{equation*}
The joint probability density for the pair $(\mathbf{c, \boldsymbol{\omega}})$ is obtained by multiplying \eqref{eq:clabel} by the prior density
\begin{equation*}
 \prior(\mathbf{c}, \boldsymbol{\omega}| \alpha, K)=\frac{\Gamma(\alpha K)}{\Gamma(\alpha)^K} \prod_{k=1}^K \omega_k^{|A_k| + \alpha - 1}.
\end{equation*}
This is still a Dirichlet probability density function of parameters ($|A_1| + \alpha, \dots, |A_K| + \alpha$) and integration with respect to $\boldsymbol{\omega}$ is straightforward
\begin{align*}
\prior(\mathbf{c}|\alpha, K)=&\frac{\Gamma(\alpha K)}{\Gamma(\alpha )^K} \int_{\boldsymbol{\omega}}\prod_{k=1}^K \omega_k^{|A_k| + \alpha - 1}d\boldsymbol{\omega}, \\
=&\frac{\Gamma(\alpha K)}{\Gamma(\alpha)^K}\frac{\prod_{k\leq K}\Gamma(|A_k| + \alpha)}{ \Gamma(\sum_{k=1}^K (|A_k| + \alpha))} \\
& \times \int_{\boldsymbol{\omega}} \text{Dir}(\boldsymbol{\omega}; |A_1| + \alpha, \dots, |A_K| + \alpha)d\boldsymbol{\omega},\\
=&\frac{\Gamma(\alpha K)}{\Gamma(\alpha)^K}\frac{\prod_{k\leq K}\Gamma(|A_k| + \alpha)}{ \Gamma(N + \alpha K)}.
\end{align*}
This integrated density corresponds to the first term on the right hand side of \eqref{eq:Dir}. The second term is obtained similarly and the joint density $\prior(\mathbf{c}, \mathbf{y} | K,D)$ follows by independence.

\section{Computational complexity}\label{sec:comp-compl}
To evaluate the computational complexity of the proposed algorithm, we assume
that the gamma function can be computed in constant time (see
\cite{numericalRecipes2007}). The core computation task consists in evaluating
the change in ICL induced by exchanges and merges. The main quantities involved
in those computations are the $(L_{kgd})_{1\leq ,k\leq g, 1\leq d\leq D}$. We
first describe how to handle those quantities and then analyze the cost of the
exchange and merge operations.

\subsection{Data structures}
The quantities $(L_{kgd})_{1\leq ,k\leq g, 1\leq d\leq D}$ are stored in a
three dimensional array that is never resized (it occupies
a $O(K_{max}^2D_{max})$ memory space) so that at any time during the
algorithm, accessing to a value or modifying it can be done in constant
time. The quantities needed to compute $L_{kgd}$, the $S_{kgd}$, $P_{kgd}$ and
$R_{kgd}$ are handled in a similar way. 

In addition, we maintain aggregated interaction counts for each time interval
and each node. More precisely, we have for instance for a time interval $I_u$ 
\begin{equation*}
S_{kgu}:=\sum_{c_i=k}\sum_{c_j=g}N^{I_u}_{ij},
\end{equation*}
and similar quantities such as $P_{kgu}$. For a node $i$, we have e.g.
\begin{equation*}
S_{igd}:=\sum_{c_j=g}\sum_{y_u=d}N^{I_u}_{ij},
\end{equation*}
and other related quantities. The memory occupied by those structures is in
$O(N^2U)$. Cluster memberships and clusters sizes are also stored in arrays. 

In order to evaluate the ICL change induced by an operation, we need to
compute its effect on $L_{kgd}$ in order to obtain $L^*_{kgd}$. This can be
done in constant time for one value. For instance moving time interval $I_u$
from $C_{d'}$ to $C_l$ implies the following modifications:
\begin{itemize}
\item $S_{kgd'}$ is reduced by $S_{kgu}$ while $S_{kgl}$ is increased by the
  same quantity;
\item $P_{kgd'}$ is divided by $P_{kgu}$ while $P_{kgl}$ is multiplied by the
  same quantity;
\item $R_{kgd'}$ is decreased by $|A_k||A_g|$ (or $|A_k|(|A_k|-1)$) while
  $R_{kgl}$  is increased by the same quantity. 
\end{itemize}

When an exchange or a fusion is actually implemented, we update all the data
structures. The update cost is dominated by the other phases of the
algorithm. For instance when $I_u$ is moved from $d'$ to $l$, we need to
update:
\begin{itemize}
\item cluster memberships and cluster sizes, which is done in $O(1)$;
\item $L_{kgd'}$ and $L_{kgl}$ for all $k$ and $g$, which is done in $O(K^2)$;
\item aggregated counts and products, such as $S_{igd'}$ and $S_{igl}$, which
  is done in $O(NKD)$. 
\end{itemize}
Considering that $K\leq N$ and $D\leq U$, the total update cost is in $O(NKD)$
for time interval related operations and in $O(UK^2)$ for node related
operations. 

\subsection{Exchanges}
The calculation of $\Delta^{E,T}_{d'\rightarrow l}$ for a time interval cluster
exchange from equation \eqref{eq:exchange} involves a sum with $K^2$
terms. As explained above each term is obtained in constant time, thus the
total computation time is in $O(K^2)$. This has to be evaluated for all time
clusters and for all time interval, summing to a total cost of $O(UDK^2)$. 

Similarly, the calculation of $\Delta^{E,V}_{d'\rightarrow l}$ involves a fix
number of sums with at most $KD$ terms in each sum. The total computation time
is therefore in $O(KD)$. This had to be evaluated for each node and for all
node clusters, summing to a total cost of $O(NK^2D)$. 

Notice that we have evaluated the total cost of one exchange round, i.e., in
the case where all time intervals (or all nodes) are considered once. This
evaluation does not take into account the reduction in the number of clusters
generally induced by exchanges. 

\subsection{Merges}
Merges are very similar to exchange in terms of computational complexity. They
involve comparable sums that can be computed efficiently using the data
structures described above. The computational cost for one time cluster merge
round is in $O(D^2K^2)$ while it is in $O(K^3D)$ for node clusters. 

\subsection{Total cost}
The worst case complexity of one full exchange phase (with each node and each
time interval considered once) is $O((N+U)D_{max}K^2_{max})$. The worst case
complexity of one merge with mixed GM is
$O(D_{max}K^2_{max}(D_{max}+K_{max}))$ which is smaller than the previous one
for $N\geq K_{max}$ and $U\geq D_{max}$. Thus the worst case complexity of one
``iteration'' of the algorithm is $O((N+U)D_{max}K^2_{max})$.

Unfortunately, the actual complexity of the algorithm, while obviously related to this
quantity, is difficult to evaluate for two reasons. Firstly, we have no way to
estimate the number of exchanges needed in the exchange phase (apart from
bounding them with the number of possible partitions). Secondly, we observe in
practice that exchanges reduce the number of clusters, especially when
$D_{max}$ and $K_{max}$ are high (i.e. close to $U$ and $N$,
respectively). Thus the actual cost of one individual exchange reduces very
quickly during the first exchange phase leading to a vast overestimation of
its cost using the proposed bounds. As a consequence, the merge phase is also
quicker than evaluated by the bounds. 

A practical evaluation of the behaviour of the algorithm, while outside the
scope of this paper, would be very interesting to assess its potential use on
large data sets.

\section*{References}

\bibliography{mybibfile}

\end{document}